# Improving the efficacy of Deep Learning models for Heart Beat detection on heterogeneous datasets

Andrea Bizzego[1,†], Giulio Gabrieli [2,†], Michelle Jin-Yee Neoh[2] and Gianluca Esposito[1,2,3,*]

1. Department of Psychology and Cognitive Science, University of Trento, 38068 Trento, Italy; andrea.bizzego@unitn.it (A.B.)
2. Psychology Program, Nanyang Technological University, Singapore 639818, Singapore; giulio001@e.ntu.edu.sg (G.G.); michelle008@e.ntu.edu.sg (M.N.JY)
3. Lee Kong Chian School of Medicine, Nanyang Technological University, Singapore 308232, Singapore
* Correspondence: gianluca.esposito@ntu.edu.sg
† These authors contributed equally to this work.

**Abstract:** Deep Learning (DL) have greatly contributed to bioelectric signals processing, in particular to extract physiological markers. However, the efficacy and applicability of the results proposed in the literature is often constrained to the population represented by the data used to train the models. In this study, we investigate the issues related to applying a DL model on heterogeneous datasets. In particular, by focusing on heart beat detection from Electrocardiogram signals (ECG), we show that the performance of a model trained on data from healthy subjects decreases when applied to patients with cardiac conditions and to signals collected with different devices. We then evaluate the use of Transfer Learning (TL) to adapt the model to the different datasets. In particular, we show that the classification performance is improved, even with datasets with a small sample size. These results suggest that a greater effort should be made towards generalizability of DL models applied on bioelectric signals, in particular by retrieving more representative datasets.

## 1. Introduction

In medicine and other medical sciences, physiological recordings are widely employed to monitor and assess the health status of patients [1,2].

The possibility of using Machine Learning (ML) and Artificial Intelligence (AI) to automatize the extraction of physiological indicators from signals has been widely explored in recent years. Numerous studies have employed AI models on preprocessed physiological signals, for instance, to identify usable segments of pupillometry measures in infants [3], as well as ventricular hypertrophy [4], arrhythmia [5], muscle fatigue [6], and stress [7]. Employment of such techniques not only allows for a reduction in the amount of time and resources required for signal processing, but also increases the reproducibility of the process, while reducing the likelihood of human errors.

Amongst the available AI techniques, Deep Neural Networks (DNN) prove to be one of the most promising [8,9]. A family of Machine Learning methods, DNN rely on the use of modular architectures, based on multiple non-linear processing units (layers), to extract high-level patterns from data. Thanks to the hierarchical structure of the layers, DNN progressively obtain high-level features from low-level representations [10], thus transforming input data into a multi-dimensional representation, that is used to solve the classification task [11]. The adoption of Convolutional Neural Networks (CNN) in applications based on medical data (bio images and physiological signals) is rapidly growing, with a wide range of applications [12–16]. Not exclusive to image processing, DNN and CNN have also been applied successfully to the analysis of physiological signals. For instance, in a study by Wieclaw *et al.* [17], a DNN was successfully employed to design a biometric identification signal based on Electrocardiogram (ECG), while Mathews *et al.* [18] employed DNN to identifying ventricular and supraventricular ectopic beats. Moreover, Xu *et al.* [19] effectively employed a DNN to classify the type of heartbeat patterns (e.g. normal beat, arrhythmia) from raw ECG recordings. Similar procedures were employed on other physiological signals: for example, Yu and Sun [20] used a DNN to classify emotions from phasic and tonic components of the Electrodermal Activity (EDA), while Mukhopadhyay and Samui [11] employed DNN to classify limbic movements from Electro-myogram (EMG) signals.

The identification of heart beats in ECG signals is one of the main task in the clinical practice that uses physiological signals. Several methods and algorithms have been proposed [21–24], but, notwithstanding its proven efficacy, the use of DNN is still limited. Silva and colleagues [25,26] presented an approach based on the combination of a CNN model with the Pan-Tompkins algorithm, a popular QRS complex detection algorithm. Despite the high effectiveness of the model proposed by Silva *et al.* [26], one of the main problems of a inaccurate identification of beats in the ECG signals may cause excessive false alarm problem, especially in Intensive Care Units (ICU) [27–29]. One of the biggest concerns of the application of DNN in clinical practices lies in the applicability of a model to signals for which it has received little or no training. For example, a model trained on a dataset consisting of signals recorded from healthy subjects may be employed in the ICU, where patients with Arrhythmia may be present. Beat identification errors in medical settings cause False Alarms that can make the identification of underlying conditions more difficult [30,31], and can affect both patient [32] and healthcare professionals wellbeing [33]. As such, before it can be extensively used in the clinical practice, any AI-based solution has to be tested to prove its reliability on datasets from different population, and with signals from different devices.

*1.1. Aim of this study*

The aim of this work is to verify the performance of using a Convolutional Deep Neural Network trained for a ECG beat detection task on the signals of healthy subjects, when signals recorded from individuals from a clinical population are fed to the model.

## 2. Materials and Methods

*2.1. Datasets*

**Table 1.** Sample sizes of the subsets used in this study for each partition. N: number of subjects; Segments: number of segments; %BEAT: percentages of segments in the BEAT class.

| | Train | | | Test | | |
|---|---|---|---|---|---|---|
| **Dataset Name** | **N** | **Segments** | **% BEAT** | **N** | **Segments** | **% BEAT** |
| NormalSinus+LongTerm | 17 | 240,000 | 7.37 | 8 | 80,000 | 8.47 |
| Arrhythmia | 32 | 230,000 | 6.19 | 16 | 110,000 | 7.07 |
| Baseline FlexComp | 12 | 14,748 | 6.48 | 6 | 7,384 | 6.43 |
| Baseline ComfTech | 12 | 14,741 | 6.62 | 6 | 7,385 | 6.19 |
| Movement ComfTech | 12 | 14,886 | 7.33 | 6 | 7,443 | 6.68 |

Data used in this study were obtained from: three datasets from Physionet [34] and from the WCS dataset [35].

Specifically, we used the MIT-BIH Normal Sinus Rhythm Database (https://doi.org/10.13026/C2NK5R) and the MIT-BIH Long-Term ECG database (https://doi.org/10.13026/C2KS3F), including long-term ECG recordings from 18 subjects and 7 subjects respectively. The two datasets were merged to compose the NormalSinus+LongTerm subset. The MIT-BIH Arrhythmia Database [36] (https://doi.org/10.13026/C2F305), including 48 30-minutes ECG recordings from 47 subjects with clinical arrhythmias, were used to compose the Arrhythmia subset.

Finally, we used the WCS dataset [35] (https://doi.org/10.21979/N9/42BBFA), including ECG signals from 18 healthy subjects, simultaneously collected with a medical grade device (FlexComp acquisition unit, Thought Technology) and a wearable device (ComfTech HeartBand) in two experimental settings: baseline and movement, each lasting 5-minutes. In particular, we considered data collected during the baseline with the FlexComp device (Baseline FlexComp subset) and with the ComfTech device (Baseline ComfTech subset) to represent data collected with different medical grade and wearable devices during resting, i.e. where signals should show no contamination from artifacts due to body movement. Then we considered data collected during movement with the ComfTech device (Movement ComfTech subset) to represent data collected in real-life contexts, i.e. where signals are likely to be affected by movement artifacts.

Subsets were divided into two partitions, used for training (Train partition) and testing (Test partition) the network models. Partition were created by dividing the subjects in each dataset into two groups, with the Train partition including approximately 66% of the subjects, and the Test partition including the remaining 33% of the subjects. The actual number of subjects composing each partition and subset is provided in Table 1.

*2.2. Signal Processing*

All datasets used in this study included ECG signals and heart beats annotations, which were used to create and label ECG signal segments. Only the first 3600 seconds of long term recordings were considered. ECG signals were segmented using fixed length non-overlapping portion (duration 0.25 s). Each segment was assigned to one of two classes: BEAT or NO-BEAT. The BEAT class was assigned if a heart beat was located between 0.1 and 0.15 seconds from the beginning of the segment; otherwise the NO-BEAT was assigned. Finally, the segment was resampled with 1000 Hz with linear interpolation to uniform the length of the segments to 250 samples. Our segmentation method is different from the one used in [25] and results in more unbalanced classes: in our subset, only $\approx 6-7\%$ of the samples belong to the BEAT class (in [25] it was 37.5%). For the processing of the signals the pyphysio Python package was used [37].

*2.3. Network Architecture*

The network architecture used in this study aims to replicate the one proposed by Silva and colleagues [25,26] and is composed of a Convolutional Part, followed by a Fully Connected Part. The Convolutional Part is composed of 4 convolutional blocks, each composed of 4 layers for one-dimensional data: (a) Batch Normalization [38], (b) Convolution (with variable number of output channels and kernel size), (c) Rectified Linear Unit activation [39] and (d) Max Pooling (with kernel size set to 2). The four blocks differ in the number of output channels and kernel size of the Convolution layer (see Figure 1).

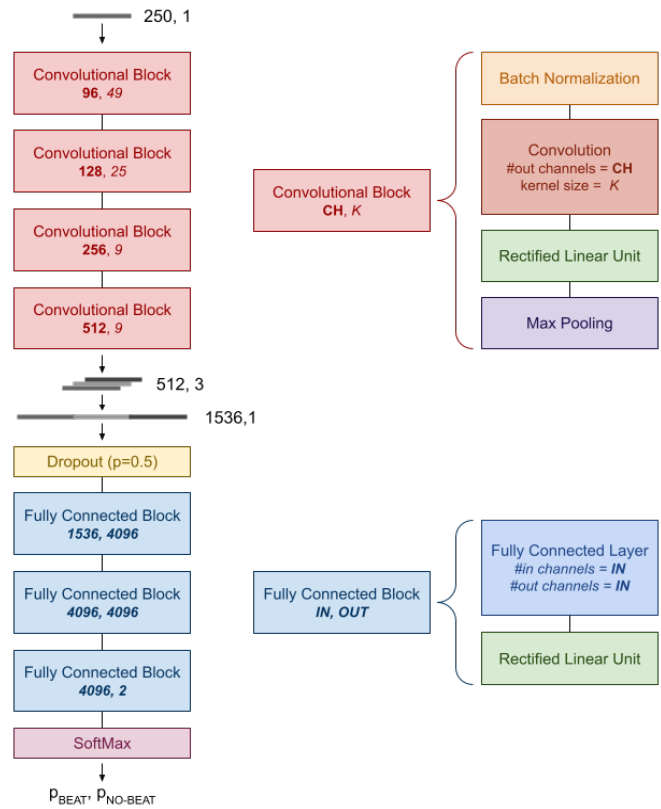

**Figure 1.** Schematic illustration of the network architecture used in this study.

The Fully Connected Part (FCP) is composed by; (a) a Dropout layer (with drop-out probability set to 0.5), (b) a sequence of 3 Fully Connected Blocks, each composed of a Fully Connected Layer and a Rectified Linear Unit activation [39], and (c) a SoftMax layer (see Figure 1).

*2.4. Network Training and Transfer Learning*

Training and testing of the network have been performed on the Gekko Cluster of the High Performance Computing Centre (HPCC, Nanyang Technological University, Singapore). To train the network, only samples in the Train partitions of the subsets are used. The training was iterated for 10 epochs, using all the segments, randomly divided into batches of 64 segments. The training is performed with back-propagation to minimize the Weighted Cross-Entropy Loss [40] between the true and predicted class. Loss weights are set to 0.06 for the NO-BEAT class and to 0.94 for the BEAT class, to compensate for the class imbalance. The AdaDelta algorithm [41] was used to optimize the network weights, with a learning rate of 0.01. The classification performance was evaluated using the Matthew Correlation Coefficient (MCC) [42] between true and predicted labels, on both the Training and Test partitions. The bootstrap technique (100 repetitions on 25% of the samples randomly selected with replacement) was used to obtain the mean MCC with 90% Confidence Intervals.

The full network is only trained once, on the NormalSinus+LongTerm subset; for the other subsets, we adapted the trained network using the transfer learning method. Specifically, we loaded the weights resulting from the training on NormalSinus+LongTerm subset and re-trained only the weights of the Fully Connected Part, while keeping the weights of the Convolutional Part.

*2.5. Experiments*

The aim of this study was to asses the performance and reproducibility of a beat detection neural network on signals from different populations and collected with different devices and in different contexts. We therefore designed three experiments, using different subsets and transfer learning:

1. *Experiment 1* - The first experiment aimed at reproducing the results of Silva and colleagues [25]. In this experiment, we trained the network using samples in the Train partition of the NormalSinus+LongTerm subsets and evaluated the performance on both the Train and Test partition of the same subset. The predictive performance was also assessed in terms of percentage of positive predicted samples (*+p*, also known as Precision), sensitivity (*Se*, also known as Recall) and F-score (*F1*), to be able to compare the results with Silva et al. study;
2. *Experiment 2* - The second experiment aimed at evaluating the performance of the trained network on the Test partition of the other subsets: (a) the Arrhythmia subset, representing a clinical population; (b) the Baseline FlexComp and (c) the Baseline ComfTech subsets, representing a normal population at rest with signals collected using another medical grade device and a wearable device respectively; (d) the Movement ComfTech subset, representing the same normal population during movement, using a wearable device;
3. *Experiment 3* - The third experiment aimed at assessing the feasibility and impact of transfer learning the trained network on the same subsets. The trained network is retrained on the Train partitions and evaluated on the Train and Test partitions.

## 3. Results

A summary of the predictive performances of the network in the three experiments, in terms of MCC with 90% CI of the prediction, are reported in Figure 2.

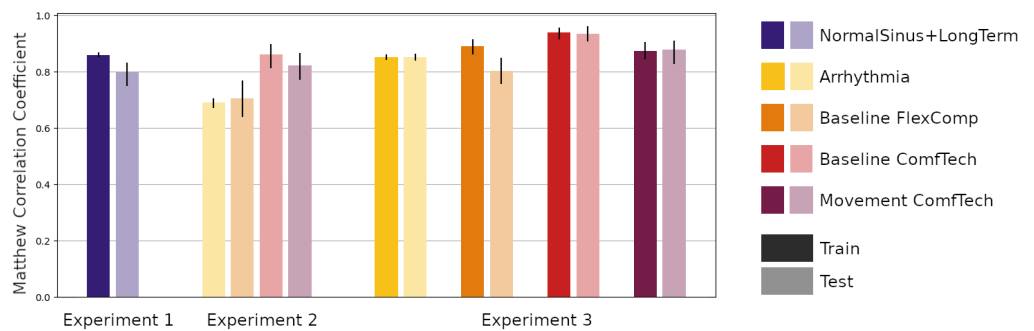

**Figure 2.** Matthew Correlation Coefficient of the networks on different datasets and partitions. Vertical bars indicate the 90% Confidence Intervals.

For what concerns the first experiment (Experiment 1, NormalSinus+LongTerm subset), the network achieved an MCC=0.860 (90% CI=[0.855, 0.866]) on the Train partition and MCC=0.797 (90% CI=[0.751, 0.830]) on the Test partition. In terms of the number of positive predicted samples, the percentage is +*p*=86.7% (90% CI=[85.9, 87.6]%) on the Train partition and +*p*=85.3% (90% CI=[81.3, 90.5]%) on the Test partition, while the measured sensitivity of the network is *Se*=87.3% (90% CI=[86.6, 88.2]%) and *Se*=78.3% (90% CI=[71.9, 82.3]%) on the Train and Test partitions, respectively. Finally, the F-score is *F1*=0.870 (90% CI=[0.864, 0.876]%) on the Train subset and *F1*=0.815 (90% CI=[0.772, 0.853]%) on the Test subset. A summary of the results in tabular form is reported in Table 2.

Table 2. Performance of the Network on the Train and Test partitions of the NormalSinus+LongTerm subset.

| Metric | Train Partition | Test Partition |
| --- | --- | --- |
| MCC | 0.860 [0.855, 0.866] | 0.797 [0.751, 0.830] |
| +p | 86.7% [85.9, 87.6] | 85.3% [81.3, 90.5] |
| Sensitivity | 87.3% [86.6, 88.2] | 78.3% [71.9, 82.3] |
| F-score | 0.870 [0.864, 0.876] | 0.815 [0.772, 0.853] |

Moving forward to the second experiment (Experiment 2), results, reported in Table 3, show that the performance of the network deteriorates when the model trained for Experiment 1 is applied to predict samples in the Arrhythmia (MCC=0.690, 90% CI=[0.675, 0.703]) and Baseline FlexComp (MCC=0.706, 90% CI=[0.642, 0.767]) subsets, while it obtained comparable results (although with larger CI) on the Baseline ComfTech (MCC=0.861, 90% CI=[0.815, 0.895]) and Movement ComfTech (MCC=0.822, 90% CI=[0.774, 0.865]) subsets.

Performing the transfer learning (Experiment 3) improves the performance of the network, although only the Fully Connected module of the network is undergoing a new training procedure. The performance of the network on the Train and Test partitions of all the subsets are similar, with the exception of the performance on the Baseline FlexComp subset. A summary of the MCC for each partition and subset, and the respective 90% CI are reported in Table 3

Table 3. Performance (MCC and 90% CI) of the Network before (Experiment 2) and after (Experiment 3) retraining; for each subset and partition under investigation.

| Dataset Name | Experiment 2 Test Partition | Experiment 3 Train Partition | Experiment 3 Test Partition |
| --- | --- | --- | --- |
| Arrhythmia | 0.690 [0.675, 0.703] | 0.852 [0.844, 0.859] | 0.852 [0.843, 0.861] |
| Baseline FlexComp | 0.706 [0.642, 0.767] | 0.852 [0.864, 0.913] | 0.803 [0.760, 0.847] |
| Baseline ComfTech | 0.861 [0.815, 0.895] | 0.939 [0.917, 0.954] | 0.935 [0.911, 0.960] |
| Movement ComfTech | 0.822 [0.774, 0.865] | 0.874 [0.846, 0.902] | 0.879 [0.830, 0.907] |

## 4. Discussion

In this paper, we conducted three experiments to investigate the performance of a Deep Learning model, trained on ECG signals collected from healthy subjects, on the identification of heart beats in signals collected from individuals from a clinical population or in signals collected with different devices.

For what concerns the first experiment (Experiment 1), using our settings, the classification performance reported in Silva *et al.* [25] could not be achieved. While the experiment reported in this work and the methods employed in Silva *et al.* [25] are similar, there are substantial differences that could explain the discrepancies in the reported sensitivity and F-score of the two models. First, the networks employed in this study and in Silva *et al.* [25] present some architectural differences, that could influence the overall results. Additionally, in this work no data augmentation procedure is performed and the analysed dataset included an unbalanced number of classes. Overall, while our network is able to identify heart beats with great accuracy, the differences between the current work and others' networks highlight the importance of replication studies to effectively assess the feasibility and applicability of different neural networks architecture for clinical tasks, and the influence of diverse modules and procedures (e.g. Data Augmentation) on the performance of the networks.

In our second experiment (Experiment 2), we verified how a network trained on the signals recorded from healthy subjects (Experiment 1) performs when employed to identify beats in segments of signals recorded from subjects with clinical conditions or on segments of signals recorded with different devices. Our results picture two different situations. Out of the four subsets of samples tested, a reduction in the performance of the network is observed in two cases (Arrhythmia and Baseline FlexComp), while similar or better performance are reported in the other two subsets (Baseline ComfTech and Movement ComfTech). One possibility for this behavior may be found in the nature of the devices used for the recording of ECG signals. Signals recorded with the ComfTech device may be less affected by noise, due to the lower sampling frequency and an internal preprocessing procedure conducted within such wearable devices.

Regarding the possibility of retraining the network on subsets of the datasets in order to explore the impact and feasibility of Transfer Learning (Experiment 3), our results indicate an increase in the performance of the network over the baseline performance (Experiment 1) for three out of the four datasets (Arrhythmia, Baseline ComfTech, Movement Comftech), while performance comparable to the baseline but superior to the one reported before transfer learning (Experiment 2) are reported for the Baseline Flexcomp dataset. Overall, the findings reported for the third experiment confirm the possible adoption of the network on segments collected from individuals with clinical conditions or on segments collected using different devices, after a re-training procedure on a subset of the dataset is conducted.

Despite the good performance of our network, the drawbacks of the application of Transfer Learning to neurophysiologcal signals should be considered. Explainability especially plays a crucial role in the adoption of a Deep Learning network, where features are not estimated prior to the analysis —as in Machine Learning studies— but are automatically learned from the network itself. As such, despite the possible higher accuracy of the model, the interpretability of the results is more complicated, especially from the biological perspective [43–46]. Generally, when the explainability of the results plays a crucial role, employing Deep Neural Networks alone is not a recommended practice. Concerning the applicability of the trained network to other domains (Transfer Learning), in order to successfully apply the model to a different problem, the network should be able to extract only generic patterns. Results of our second experiment, reported in Figure 2, show that the model's performance are lower when employed to classify data collected from clinical situation or devices for which it has not been trained, suggesting that the model is learning more than generic patterns. The computational effort required to adapt the network to a different kind of signal —ECG of individuals with clinical conditions, or ECG collected with a different device, in the current paper— was measured in Experiment 3. After being retrained with the addition of a portion of data from a new dataset, the performance of the network when employed to classify different signals improved and in some cases outperformed the original baseline (Experiment 1). These results suggest that when the model is trained on a multivariate dataset, its application with a Transfer Learning approach is more feasible. This is crucial especially for the possible application of Deep Neural Networks in a clinical setting, where high accuracy of the models is required, and for which the transparency and reproducibility of the results is essential [47]. While the clinical field may benefit from the adoption of Machine Learning models, whose possible application for the medical and research fields has been explored with regards to different types of signals and clinical situations [3,35,48–51], it is important to also consider the time and resources needed to train and deploy such models, as compared to more simplistic heuristics of linear algorithms. Future studies should consider testing the network here proposed on different datasets, and reporting the performances both before and after transfer learning. Moreover, future works should compare the computational workload and performance of simpler heuristics of algorithms with more complex Machine Learning or Neural Network models.

## 5. Conclusions

In this paper, we presented three different experiments on the application of a Deep Learning model to the identification of heart beats in ECG signals. Our results confirm the possibility of successfully employing a DNN to identify beats in ECG signals recorded from individuals with different medical backgrounds or collected with devices of different nature (e.g. clinical or wearables), by adopting a transfer learning procedure that only retrains a section of the network on a subset of a different dataset.

**Author Contributions:** Conceptualization, A.B., G.G. and G.E.; methodology, software, validation, and data curation, A.B. and G.G.; writing—original draft preparation, A.B. and G.G.; writing—review and editing, A.B., G.G., M.N.JY.; supervision, G.E.; All authors have read and agreed to the published version of the manuscript.

**Funding:** A.B. was supported by a Post-doctoral Fellowship within MIUR programme framework "Dipartimenti di Eccellenza" (DiPSCO, University of Trento).

**Institutional Review Board Statement:** Not applicable.

**Informed Consent Statement:** Informed consent was obtained from all subjects involved in the original studies.

**Data Availability Statement:** Data used in this study are derived from public datasets. They can be retrieved usink the DOI links provided in the Datasets subsection (2.1). Scripts to create the subsets, train the models and reproduce the results are publicly available at: https://gitlab.com/abp-san-public/dl-beat-detection

**Conflicts of Interest:** The authors declare no conflict of interest.

## References

1. Wagner, J.; Kim, J.; André, E. From physiological signals to emotions: Implementing and comparing selected methods for feature extraction and classification. 2005 IEEE international conference on multimedia and expo. IEEE, 2005, pp. 940–943.
2. Gabrieli, G.; Azhari, A.; Esposito, G. PySiology: A python package for physiological feature extraction. In *Neural Approaches to Dynamics of Signal Exchanges*; Springer, 2020; pp. 395–402.
3. Gabrieli, G.; Balagtas, J.P.M.; Esposito, G.; Setoh, P. A Machine Learning approach for the automatic estimation of fixation-time data signals' quality. *Sensors* **2020**, *20*, 6775.
4. Jothiramalingam, R.; Jude, A.; Patan, R.; Ramachandran, M.; Duraisamy, J.H.; Gandomi, A.H. Machine learning-based left ventricular hypertrophy detection using multi-lead ECG signal. *Neural Computing and Applications* **2020**, pp. 1–11.
5. Bulbul, H.I.; Usta, N.; Yildiz, M. Classification of ECG arrhythmia with machine learning techniques. 2017 16th IEEE International Conference on Machine Learning and Applications (ICMLA). IEEE, 2017, pp. 546–549.
6. Karthick, P.; Ghosh, D.M.; Ramakrishnan, S. Surface electromyography based muscle fatigue detection using high-resolution time-frequency methods and machine learning algorithms. *Computer Methods and Programs in Biomedicine* **2018**, *154*, 45–56.
7. Zontone, P.; Affanni, A.; Bernardini, R.; Piras, A.; Rinaldo, R. Stress detection through electrodermal activity (EDA) and electrocardiogram (ECG) analysis in car drivers. 2019 27th European Signal Processing Conference (EUSIPCO). IEEE, 2019, pp. 1–5.
8. Manzalini, A. Towards a Quantum Field Theory for Optical Artificial Intelligence. *Annals of Emerging Technologies in Computing (AETiC), Print ISSN* **2019**, pp. 2516–0281.
9. Sánchez-Sánchez, C.; Izzo, D.; Hennes, D. Learning the optimal state-feedback using deep networks. 2016 IEEE Symposium Series on Computational Intelligence (SSCI). IEEE, 2016, pp. 1–8.
10. LeCun, Y.; Bengio, Y.; Hinton, G. Deep Learning. *Nature* **2015**, *521*, 436–444.
11. Mukhopadhyay, A.K.; Samui, S. An experimental study on upper limb position invariant EMG signal classification based on deep neural network. *Biomedical Signal Processing and Control* **2020**, *55*, 101669.
12. Bizzego, A.; Bussola, N.; Salvalai, D.; Chierici, M.; Maggio, V.; Jurman, G.; Furlanello, C. Integrating deep and radiomics features in cancer bioimaging. 2019 IEEE Conference on Computational Intelligence in Bioinformatics and Computational Biology (CIBCB). IEEE, 2019, pp. 1–8.
13. Tseng, H.H.; Wei, L.; Cui, S.; Luo, Y.; Ten Haken, R.K.; El Naqa, I. Machine learning and imaging informatics in Oncology. *Oncology* **2018**, *Nov 23*, 1–19.
14. Topol, E.J. High-performance medicine: The convergence of human and artificial intelligence. *Nature Medicine* **2019**, *25*, 44.
15. Esteva, A.; Kuprel, B.; Novoa, R.A.; Ko, J.; Swetter, S.M.; Blau, H.M.; Thrun, S. Dermatologist-level classification of skin cancer with deep neural networks. *Nature* **2017**, *542*, 115–118.
16. Mobadersany, P.; Yousefi, S.; Amgad, M.; Gutman, D.A.; Barnholtz-Sloan, J.S.; Velázquez Vega, J.E.; Brat, D.J.; Cooper, L.A.D. Predicting cancer outcomes from histology and genomics using convolutional networks. *Proceedings of the National Academy of Sciences* **2018**, *115*, E2970–E2979.

17. Wieclaw, L.; Khoma, Y.; Fałat, P.; Sabodashko, D.; Herasymenko, V. Biometric identification from raw ECG signal using deep learning techniques. 2017 9th IEEE International Conference on Intelligent Data Acquisition and Advanced Computing Systems: Technology and Applications (IDAACS). IEEE, 2017, Vol. 1, pp. 129–133.
18. Mathews, S.M.; Kambhamettu, C.; Barner, K.E. A novel application of deep learning for single-lead ECG classification. *Computers in Biology and Medicine* **2018**, *99*, 53–62.
19. Xu, S.S.; Mak, M.W.; Cheung, C.C. Towards end-to-end ECG classification with raw signal extraction and deep neural networks. *IEEE Journal of Biomedical and Health Informatics* **2018**, *23*, 1574–1584.
20. Yu, D.; Sun, S. A systematic exploration of deep neural networks for EDA-based emotion recognition. *Information* **2020**, *11*, 212.
21. Li, Q.; Mark, R.G.; Clifford, G.D. Robust heart rate estimation from multiple asynchronous noisy sources using signal quality indices and a Kalman filter. *Physiological measurement* **2007**, *29*, 15.
22. Tarassenko, L.; Townsend, N.; Clifford, G.; Mason, L.; Burton, J.; Price, J. Medical signal processing using the software monitor **2001**.
23. Kohler, B.U.; Hennig, C.; Orglmeister, R. The principles of software QRS detection. *IEEE Engineering in Medicine and biology Magazine* **2002**, *21*, 42–57.
24. Ebrahim, M.H.; Feldman, J.M.; Bar-Kana, I. A robust sensor fusion method for heart rate estimation. *Journal of clinical monitoring* **1997**, *13*, 385–393.
25. Silva, P.; Luz, E.; Wanner, E.; Menotti, D.; Moreira, G. QRS detection in ECG signal with convolutional network. Iberoamerican Congress on Pattern Recognition. Springer, 2018, pp. 802–809.
26. Silva, P.; Luz, E.; Silva, G.; Moreira, G.; Wanner, E.; Vidal, F.; Menotti, D. Towards better heartbeat segmentation with deep learning classification. *Scientific Reports* **2020**, *10*, 1–13.
27. Chambrin, M.C. Alarms in the intensive care unit: how can the number of false alarms be reduced? *Critical Care* **2001**, *5*, 1–5.
28. Eerikäinen, L.M.; Vanschoren, J.; Rooijakkers, M.J.; Vullings, R.; Aarts, R.M. Reduction of false arrhythmia alarms using signal selection and machine learning. *Physiological measurement* **2016**, *37*, 1204.
29. Gal, H.; Liel, C.; O'Connor Michael, F.; Idit, M.; Lerner, B.; Yuval, B. Machine learning applied to multi-sensor information to reduce false alarm rate in the ICU. *Journal of clinical monitoring and computing* **2020**, *34*, 339–352.
30. Sendelbach, S.; Funk, M. Alarm fatigue: a patient safety concern. *AACN advanced critical care* **2013**, *24*, 378–386.
31. Drew, B.J.; Harris, P.; Zègre-Hemsey, J.K.; Mammone, T.; Schindler, D.; Salas-Boni, R.; Bai, Y.; Tinoco, A.; Ding, Q.; Hu, X. Insights into the problem of alarm fatigue with physiologic monitor devices: a comprehensive observational study of consecutive intensive care unit patients. *PloS one* **2014**, *9*, e110274.
32. Xie, H.; Kang, J.; Mills, G.H. Clinical review: The impact of noise on patients' sleep and the effectiveness of noise reduction strategies in intensive care units. *Critical Care* **2009**, *13*, 1–8.
33. Sorkin, R.D. Why are people turning off our alarms? *The Journal of the Acoustical Society of America* **1988**, *84*, 1107–1108.
34. Goldberger, A.L.; Amaral, L.A.; Glass, L.; Hausdorff, J.M.; Ivanov, P.C.; Mark, R.G.; Mietus, J.E.; Moody, G.B.; Peng, C.K.; Stanley, H.E. PhysioBank, PhysioToolkit, and PhysioNet: components of a new research resource for complex physiologic signals. *circulation* **2000**, *101*, e215–e220.
35. Bizzego, A.; Gabrieli, G.; Furlanello, C.; Esposito, G. Comparison of wearable and clinical devices for acquisition of peripheral nervous system signals. *Sensors* **2020**, *20*, 6778.
36. Moody, G.B.; Mark, R.G. The impact of the MIT-BIH arrhythmia database. *IEEE Engineering in Medicine and Biology Magazine* **2001**, *20*, 45–50.
37. Bizzego, A.; Battisti, A.; Gabrieli, G.; Esposito, G.; Furlanello, C. pyphysio: A physiological signal processing library for data science approaches in physiology. *SoftwareX* **2019**, *10*, 100287.
38. Ioffe, S.; Szegedy, C. Batch normalization: Accelerating deep network training by reducing internal covariate shift. *arXiv preprint arXiv:1502.03167* **2015**.
39. Shang, W.; Sohn, K.; Almeida, D.; Lee, H. Understanding and improving convolutional neural networks via concatenated rectified linear units. International Conference on Machine Learning, 2016, pp. 2217–2225.
40. Zhang, Z.; Sabuncu, M. Generalized cross entropy loss for training deep neural networks with noisy labels. *Advances in Neural Information Processing Systems* **2018**, *31*, 8778–8788.
41. Zeiler, M.D. Adadelta: an adaptive learning rate method. *arXiv preprint arXiv:1212.5701* **2012**.
42. Jurman, G.; Riccadonna, S.; Furlanello, C. A comparison of MCC and CEN error measures in multi-class prediction. *PloS ONE* **2012**, *7*, e41882.
43. Sun, W.; Zheng, B.; Qian, W. Automatic feature learning using multichannel ROI based on deep structured algorithms for computerized lung cancer diagnosis. *Computers in Biology and Medicine* **2017**, *89*, 530–539.
44. Arimura, H.; Soufi, M.; Kamezawa, H.; Ninomiya, K.; Yamada, M. Radiomics with artificial intelligence for precision medicine in radiation therapy. *Journal of Radiation Research* **2019**, *60*, 150–157.
45. Li, Z.; Wang, Y.; Yu, J.; Guo, Y.; Cao, W. Deep learning based radiomics (DLR) and its usage in noninvasive IDH1 prediction for low grade glioma. *Scientific reports* **2017**, *7*, 1–11.
46. Kontos, D.; Summers, R.M.; Giger, M. Special section guest editorial: Radiomics and deep learning. *Journal of Medical Imaging* **2017**, *4*.

47. Haibe-Kains, B.; Adam, G.A.; Hosny, A.; Khodakarami, F.; Waldron, L.; Wang, B.; McIntosh, C.; Goldenberg, A.; Kundaje, A.; Greene, C.S.; others. Transparency and reproducibility in artificial intelligence. *Nature* **2020**, *586*, E14–E16.
48. Gabrieli, G.; Bizzego, A.; Neoh, M.J.Y.; Esposito, G. fNIRS-QC: Crowd-Sourced Creation of a Dataset and Machine Learning Model for fNIRS Quality Control. *Applied Sciences* **2021**, *11*, 9531.
49. Bizzego, A.; Gabrieli, G.; Esposito, G. Deep Neural Networks and Transfer Learning on a Multivariate Physiological Signal Dataset. *Bioengineering* **2021**, *8*, 35.
50. Bizzego, A.; Gabrieli, G.; Azhari, A.; Setoh, P.; Esposito, G. Computational methods for the assessment of empathic synchrony. In *Progresses in Artificial Intelligence and Neural Systems*; Springer, 2021; pp. 555–564.
51. Gabrieli, G.; Bornstein, M.H.; Manian, N.; Esposito, G. Assessing Mothers' Postpartum Depression From Their Infants' Cry Vocalizations. *Behavioral Sciences* **2020**, *10*, 55.